\documentclass[runningheads]{llncs}

\usepackage[utf8]{inputenc}
\usepackage{amssymb}
\usepackage{stmaryrd}
\usepackage{amsmath}
\usepackage{booktabs}
\usepackage{verbatim}
\usepackage{hyperref}
\usepackage{tikz}
\usetikzlibrary{graphs,quotes,positioning,shapes,matrix}
\usetikzlibrary{arrows.meta,decorations,er,calc,fit}
\tikzset{
  msg/.style={fill=white, draw, solid, font=\footnotesize\ttfamily,
    thin, align=left, near start, inner sep=2pt},
  send/.style={->, densely dotted, thick}
}

\newtheorem{competency}{CQ}

\makeatletter
\def\verbatim@font{\footnotesize\ttfamily}
\makeatother

\title{Towards an ontology of HTTP interactions}

\author{Mathieu Lirzin\inst{1,2}\orcidID{0000-0002-8366-1861} \and\\
  Béatrice Markhoff\inst{2}\orcidID{0000-0002-5171-8499}}
\authorrunning{M. Lirzin and B. Markhoff}

\institute{
  Néréide, 8 rue des déportés, 37000 Tours, France\\
  \email{mathieu.lirzin@nereide.fr} 
  \and 
  \textsc{LIFAT EA 6300}, Université de Tours, Tours, France\\
  \email{beatrice.markhoff@univ-tours.fr} 
}

\begin{document}
\maketitle
\thispagestyle{empty}

\begin{abstract}
Enterprise information systems have adopted Web-based foundations for 
exchanges between heterogeneous programmes. These programs provide 
and consume via Web APIs some resources identified by URIs, 
whose representations are transmitted via HTTP. Furthermore HTTP remains 
at the heart of all Web developments (Semantic Web, linked data, IoT...). 
Thus, situations where a program must be able to reason about HTTP interactions 
(request-response) are multiplying. This requires 
an explicit formal specification of a shared conceptualization of those interactions. 
A proposal for an RDF vocabulary exists, 
developed with a view to carrying out web application conformity tests 
 and record the tests outputs. This vocabulary has already been reused. 
 In this report we propose to adapt and extend it for making it more reusable. \\
 
\textbf{The content of this report has been published in French \cite{lirzin2020}\footnote{\url{https://hal.archives-ouvertes.fr/hal-02888065}} at IC 2020.}

\end{abstract}

\section{Introduction}
\label{sec:intro}
The Hypertext Transfer Protocol (HTTP) is, together with the logical
addressing (URI) and HTML, the basic building block of the Web.  While
HTTP remains at the heart of Web development, current enterprise
information systems use it routinely for exchanges between
heterogeneous programs: to this aim, all is needed is an HTTP server
on the one side and an HTTP client on the other. Despite its apparent
simplicity the HTTP protocol is quite large, it covers all aspects of
client-server communications, while keeping it open for potential
evolutions. The size of the HTTP 1.1 specification which consists
currently in 8 IETF RFCs, refined and extended by other RFCs,
demonstrates that fact.  This specification defines constraints that
impact not only the implementers of Web Servers but also application
developers that are implementing request handlers: they need to ensure
that their request handlers conform to the semantics of HTTP.  Our
goal is to formalize the HTTP specification to be able to describe
\textit{Web interactions} in a sufficiently precise way for enabling a
mechanized verification of protocol conformance.  Since the beginning
of the Web various proposals have been made to describe interactions
between Web clients and a Web server, with different means and
objectives. Many of them are "machine-readable" but are limited to the
syntactic level and do not rely on an ontological description of HTTP
interactions.  The main difficulty when describing Web interactions
consists in taking into account the presence of hypermedia links in
the representation of the Web resources. The point is that HTTP mixes
the data and the control over that data. This makes things easy for
humans (developers) to interact with.  However it is hard to describe
it formally, and it is still difficult for machines to interact
robustly when using it.  In this paper we propose an ontology
specified in Description Logic for describing Web interactions, with
the following contributions:
\begin{itemize}
\item We use the
  $\mathcal{SROIQ^{(D)}}$ Description Logic to describe the proposed
   ontology, which refines the initial
 W3C draft RDF Vocabulary for HTTP~\cite{koch_http_2017}. 
 In this way we characterize more precisely the
  properties between already identified elements and we properly introduce 
   the new ones.
\item HTTP message headers represent an associative array with
  heterogeneous value types: we provide a way to describe headers in
  a generic manner while enabling to express more precisions for specific ones
  such as the \verb=Location= header whose value is linked to a new
  resource (Section \ref{sec:headers}).
\item HTTP allows the usage of various representation formats in the
  body payload by making use of Media Type declarations. This variability 
  is required for HTTP genericity, but it makes it
  difficult to formalize the interaction aspect of the protocol regarding
  the exchanged content: we propose a way to describe uniformly both the data 
  and meta-data of messages for a subset of the allowed content types
  (Section \ref{sec:body}).
\item URIs primary goal is to identify a resource, however in the context
  of Web APIs the optional query part of the URI is extensively used to
  parameterize the behavior of the server request handler. We define a
  set of properties that can be used to represent the various parts of an
  URI and show how query parameters can be accessed
  (Section~\ref{sec:query-params}).
\item Our ontology is implemented\footnote{\url{https://labs.nereide.fr/mthl/http}} with Protégé using
  OWL 2 DL, which
  corresponds to the $\mathcal{SROIQ^{(D)}}$ Description Logic with the
  specificity of using URIs as identifiers for both classes and individuals. 
  This facilitates the description of the hypermedia links present in both the
  message data and meta-data.
  We use the HermiT reasoner to validate its satisfiability and
  consistency. We also added some representative sets of
  individuals, representing real Web interactions, to verify that 
  the SPARQL queries implementing our Competency Questions 
  (representing our knowledge representation needs) can be  performed.
\end{itemize}
We first provide some background on the HTTP protocol with an overview
of the W3C RDF vocabulary for HTTP in Section~\ref{sec:background}, we
precise our problem requirements in Section~\ref{sec:pbstate}, and we 
present our ontology in Section~\ref{sec:ontology}.  Before concluding
in Section~\ref{sec:conclusion} we situate our proposal with respect
to existing solutions that tackled the issue of describing HTTP
interactions in Section~\ref{sec:related}.


\section{Background}
\label{sec:background}
HTTP is a standard managed by the Internet Engineering Task Force
(IETF).  Starting from the first proposal of Tim Berners-Lee, a way to
describe the transmitted data formats (MIME headers) was early
integrated, then came many other efficiency-related features
(persistent connection, caching) together with transmission security
considerations (HTTPs, encryption), which result in the power of
expressing rich and diverse information about client-server interactions.  The
last HTTP 1.1 specification, produced in 2014, is declined in a serie of 8
RFCs where the principal ones are RFC~7230 which defines the message
syntax and routing aspect of the protocol and RFC~7231 which defines
the semantics and content aspects.  Despite the clarifications
contained in these RFC, and the proposed optimisations in HTTP
2~\footnote{\url{https://tools.ietf.org/html/rfc7540}}, this standard on
which a vast majority of current applications are based today stays
voluntarily open to make room for new inventions.

While the consultation of the IETF RFCs is always useful and necessary
for developers, programs also regularly have to deal with elements of
this protocol and would benefit from being able to interpret it, at
least partly.  An RDF vocabulary for HTTP was developed to support Web
accessibility evaluation tools~\cite{koch_http_2017}.  It is used in
EARL~\footnote{\url{https://www.w3.org/WAI/standards-guidelines/earl/}}, a
format for expressing the evaluation results, and describes
essentially the HTTP headers exchanged between a client and a
server. Although complete for their requirements, the authors left it
at the stage of a proposal (W3C Working Group Note, 2017).

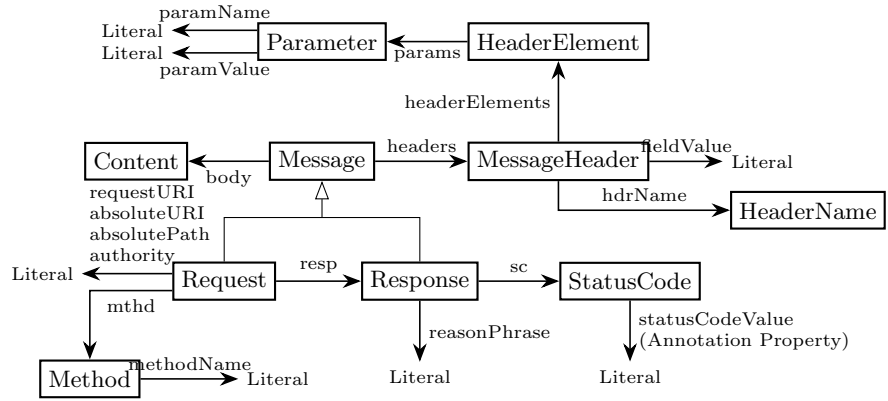
\begin{figure}
  \centering
  \begin{tikzpicture}
    \tikzset {
      class/.style={draw, solid, thick, font=\footnotesize,
        minimum height=1.5em, minimum width=4em, node distance=3.3em},
      lit/.style={font=\scriptsize, node contents=Literal, node distance=2.5em},
      prop/.style={arrows={-Stealth[length=7pt]}, semithick},
      lbl/.style={auto, font=\scriptsize}
    };
    \node[class] (Param) {Parameter};
    \node[class] (HElem) [right=of Param] {HeaderElement};
    \node[class] (MHdr) [below=of HElem] {MessageHeader};
    \node[class] (HN) [right=of MHdr, yshift=-2em] {HeaderName};
    \node[class] (Msg) [below=of Param] {Message};
    \node[coordinate] (MsgB) [below=1.5em of Msg] {};
    \node[class] (Cnt) [left=of Msg] {Content};
    \node[class] (Req) [below=of Msg, xshift=-4em] {Request};
    \node[class] (Mthd) [left=of Req, xshift=2em, yshift=-4em] {Method};
    \node[class] (Resp) [below=of Msg, xshift=4em] {Response};
    \node[class] (SC) [right=of Resp] {StatusCode};

    \node (ParamT) [coordinate, above=1.5mm of Param.west] {};
    \node (ParamB) [coordinate, below=1.5mm of Param.west] {};
    \node (pN) [lit, left=of ParamT, xshift=-1em];
    \node (pV) [lit, left=of ParamB, xshift=-1em];
    \draw[prop] (ParamT) to node[lbl, above] {paramName} (pN);
    \draw[prop] (ParamB) to node[lbl, below] {paramValue} (pV);

    \node (rP) [lit, below=of Resp];
    \draw[prop] (Resp) to node[lbl] {reasonPhrase} (rP);

    \node (mN) [lit, right=of Mthd, xshift=1.5em];
    \draw[prop] (Mthd) to node[lbl] {methodName} (mN);

    \node (ReqT) [coordinate, above=1mm of Req.west] {};
    \node (uris) [lit, left=of ReqT, xshift=-1.2em];

    \node (scv) [lit, below=of SC];
    \draw[prop] (SC) -- node[lbl, align=left] {
      statusCodeValue \\
      (Annotation Property)
    } (scv);

    \draw[prop] (ReqT) -- node[auto, above, font=\scriptsize,
      align=left, near start]{
      requestURI \\
      absoluteURI \\
      absolutePath \\
      authority
    } (uris);

    \draw (Req) |- (MsgB);
    \draw (Resp) |- (MsgB);
    \draw[arrows = {-Stealth[inset=0em, fill=white,length=7pt]}] (MsgB) -- (Msg);

    \node (fV) [lit, right=3em of MHdr];
    \draw[prop] (MHdr) -- node[lbl] {fieldValue} (fV);

    \draw[prop] (HElem) to node[lbl] {params} (Param);
    \draw[prop] (MHdr) to node[lbl] {headerElements} (HElem);
    \draw[prop] (MHdr) |- node[lbl, near end] {hdrName} (HN);
    \draw[prop] (Msg) to node[lbl] {headers} (MHdr);
    \draw[prop] (Msg) to node[lbl] {body} (Cnt);
    \draw[prop] (Req) to node[lbl] {resp} (Resp);
    \draw[prop] (Req) [yshift=-1em] -| node[lbl,near start, below] {mthd} (Mthd);
    \draw[prop] (Resp) to node[lbl] {sc} (SC);
  \end{tikzpicture}
  \caption{Main elements of W3C HTTP RDF Vocabulary.}
  \label{fig:rdf-vocab}
\end{figure}

This proposal can be considered as an application ontology~\cite{Guarino1998}, 
whose purpose is to represent specificities of HTTP interactions on which
Web APIs rely, while a related 
domain ontology would allow, for instance, to declare and describe
any specific problem, function, or algorithm, whether executed using Web APIs or not. 
Figure~\ref{fig:rdf-vocab} gives a visual representation
of main classes and properties contained in the HTTP RDF 
Vocabulary~\cite{koch_http_2017}. 
This vocabulary defines 14 classes and 25 properties in the 
\verb=http= namespace, plus 11 classes and many properties
(including those of DCterm vocabulary) in four other namespaces, 
dedicated to content, headers, methods and status codes descriptions.

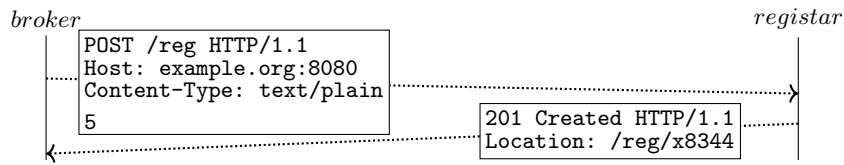
\begin{figure}
  \centering
  \begin{tikzpicture}[node distance=7.5cm]
    \node (client) at (0, 0) {$broker$};
    \node (server) at (10, 0) {$registar$};
    \node[below of=server, node distance=2cm] (server-ground) {};
    \node[below of=client, node distance=2cm] (client-ground) {};
    \draw (client) -- (client-ground);
    \draw (server) -- (server-ground);
    \draw[send] ($(client)!0.4!(client-ground)$) -- ($(server)!0.5!(server-ground)$)
    node[msg] {
      POST /reg HTTP/1.1 \\[-0.25em]
      Host: example.org:8080 \\[-0.25em]
      Content-Type: text/plain \\[0.1em]
      5
    };
    \draw[send] ($(server)!0.7!(server-ground)$) -- ($(client)!0.9!(client-ground)$)
    node[msg] {
      201 Created HTTP/1.1 \\[-0.25em]
      Location: /reg/x8344
    };
  \end{tikzpicture}
  \caption{Interaction between a broker client and a registar server}
  \label{fig:reg-inter}
\end{figure}

Basically HTTP is a request/response message protocol where a client
sends a request message to a server which then replies with a response
message. Architecturally things are more complex because there are
intermediaries (proxy/gateway) which intercept messages to provide
for example the possibility to forward messages or to implement some
caching mechanism.  However from the point of view of a client agent
those details do not impact the simple request/response model. As
illustrated in Figure~\ref{fig:rdf-vocab}, a message has a header part
and a body part and there is two kinds of messages, the requests and
the responses. A request is further characterized by a URI and a
method, and a response comes with a status code.  Let us use a simple
example to explain how to represent a Web API interaction with this
vocabulary.  This example represented in Figure~\ref{fig:reg-inter}
consists in an interaction between a broker client agent sending a
request to a registar server for registering a number of identifiers.
The request contains the wanted number of identifiers to register and
the server replies with a 201 status code to denote the creation of a
new resource corresponding to that registered collection of
identifiers.  We provide in Figure~\ref{fig:reg-inter-ttl} the RDF
representation of this interaction, following W3C HTTP RDF Vocabulary.
The used RDF serialisation is Turtle with namespace prefixes reported
in Table~\ref{table:ns}.  The RDF graph consists in a pair of
instances of \verb=http:Request= and \verb=http:Response= classes
which are linked by a \verb=http:resp= property, which defines one
interaction.  An instance of \verb=http:Request= must have a method
and a target URI.  In our example the property used for the target URI
is \verb=http:absolutePath=, a sub-property of
\verb=http:requestURI=. Notice that for the header content we use the
\verb=hdrName= / \verb=fieldValue= representation because in this
example header names are predefined ones, while it is also possible to
describe any list of key-value pairs by using class\verb= Parameter=.

\begin{figure}
  \centering
\begin{verbatim}
_:req a http:Request ;
    http:mthd mthd:POST ;
    http:absolutePath "/reg" ;
    http:headers [
        http:hdrName hdr:Host ;
        http:fieldValue "example.org:8080"
        ] , [
        http:hdrName hdr:ContentType ;
        http:fieldValue "text/plain"
        ];
    http:resp _:resp ;
    http:body [
        a cnt:ContentAsBase64 ;
        dct:isFormatOf [
            a cnt:ContentAsText ;
            cnt:chars "5"
            ]
        ] .

_:resp a http:Response ;
    http:sc sc:Created ;
    http:headers [
        http:hdrName hdr:Location ;
        http:fieldValue "/reg/x8344"
        ] .
\end{verbatim}
  \caption{Turtle representation of the registar interaction example}
  \label{fig:reg-inter-ttl}
\end{figure}

\begin{table}
  \centering
  \caption{Namespace prefixes used in the registar examples}
  \begin{tabular}{@{}ll@{}}
    \toprule
    \textbf{Prefix} & \textbf{Namespace} \\
    \midrule
    http: & \verb+http://www.w3.org/2011/http#+ \\
    mthd: & \verb+https://www.w3.org/2011/http-methods#+ \\
    hdr: & \verb+http://www.w3.org/2011/http-headers#+ \\
    sc: & \verb+http://www.w3.org/2011/http-statusCodes#+ \\
    cnt: & \verb+http://www.w3.org/2011/content#+ \\
    dct: & \verb+http://purl.org/dc/terms/+ \\
    \bottomrule
  \end{tabular}
  \label{table:ns}
\end{table}

An important aspect of the W3C HTTP RDF Vocabulary is that it reifies
the methods, headers and status codes whose semantics is precisely
defined in RFCs, while also representing them by a string literal
which enables to deal with ad-hoc headers or status codes. This
matches the extensibility requirement of the HTTP protocol. In the
registar example of Figure~\ref{fig:reg-inter}, we are only using
standard methods, headers and status code. As a consequence they are
identified by a URI instead of a literal.  An interaction is an
instance of property \verb=http:resp=. It is characterized by the
status code number associated to the response and the proposed
vocabulary allows user to determine it by identifying the class
associated with this status code number.  The status code number must
have 3 digits and its class is defined by the first digit where:
\begin{center}
  \begin{tabular}{@{}rlllll@{}}
    \toprule
    \textbf{Class} & Informational & Successful & Redirection & Client Error & Server Error \\
    \midrule
    \textbf{Status codes} &
    $\llbracket 100, 199 \rrbracket$ & $\llbracket 200, 299
    \rrbracket$ & $\llbracket 300, 399 \rrbracket$ & $\llbracket 400,
    499 \rrbracket$ & $\llbracket 500, 599 \rrbracket$ \\
    \bottomrule
  \end{tabular}
\end{center}
Those classes are defined in the namespace associated with the prefix
\verb=sc:= which is part of the HTTP vocabulary. Each reified status
code is an instance of its corresponding class. In our example, the
status code is 201 and we represent it by the individual
\verb=sc:Created=, instance of class \verb=sc:Successful=.

Another aspect which is important to consider is the body part of the
message, which is its content payload. The HTTP protocol supports the
usage of multiple formats for the same resource, which are identified
by Media-Type defined by the \verb=Content-Type= header.  In our
example the request contains the literal "5" with Media-Type
\verb=text/plain=.  The \verb=cnt:Content= class provides a way to
associate multiple representation views of the same body payload.
The \verb=http:body= property is always associated with a
\verb=cnt:ContentAsBase64= but we can associate it with the
\verb=dct:hasFormat= property to a textual representation using a
\verb=cnt:ContentAsText= resource.


\section{Problem Statement}
\label{sec:pbstate}
Other works use this RDF vocabulary, for example~\cite{VerborghAHRMSG17} uses it to define RESTdesc, a hypermedia API
definition framework for automatic composition of hypermedia
APIs. "Hypermedia API" refers to those Web APIs that effectively
follow the four constraints of the Representational State Transfer
(REST) architectural style \cite{gfielding2002principled}, in
particular the fourth: \emph{hypermedia as the engine of application
  state}.  We are also interested in hypermedia APIs, with a goal of
automatic verification of the conformance of a client requirement
 with respect to a server supply: we aim at representing both the
 requirement and the supply specifications as RDF graphs.
 This brings us to the need to formally
represent the interaction between an HTTP client and an HTTP server,
both of them using hypermedia with RDF linked content.  But we found
limitations to simply use the proposition from~\cite{koch_http_2017}
as~\cite{VerborghAHRMSG17} does.

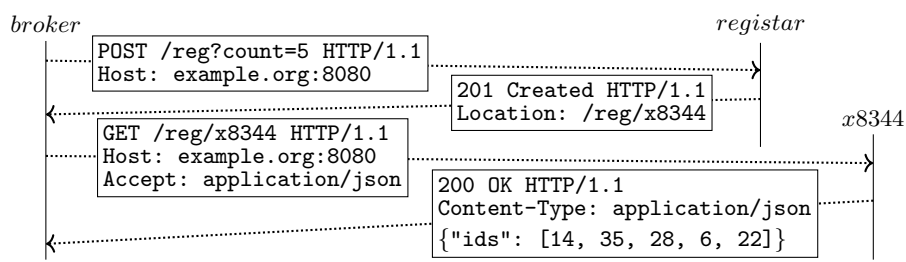
\begin{figure}
  \centering
  \begin{tikzpicture}
    \node (client) at (0, 0) {$broker$};
    \node (server) at (9.5, 0) {$registar$};
    \node (server2) at (11, -1.25) {$x8344$};
    \node[below of=client, node distance=3.25cm] (client-ground) {};
    \node[below of=server, node distance=1.75cm] (server-ground) {};
    \node[below of=server2, node distance=2cm] (server2-ground) {};
    \draw (client) -- (client-ground);
    \draw (server) -- (server-ground);
    \draw (server2) -- (server2-ground);
    \draw[send] ($(client)!0.15!(client-ground)$) -- ($(server)!0.35!(server-ground)$)
    node[msg, pos=0.3] {
      POST /reg?count=5 HTTP/1.1 \\[-0.25em]
      Host: example.org:8080
    };
    \draw[send] ($(server)!0.57!(server-ground)$) -- ($(client)!0.37!(client-ground)$)
    node[msg] {
      201 Created HTTP/1.1 \\[-0.25em]
      Location: /reg/x8344
    };
    \draw[send] ($(client)!0.54!(client-ground)$) -- ($(server2)!0.3!(server2-ground)$)
    node[msg] {
      GET /reg/x8344 HTTP/1.1 \\[-0.25em]
      Host: example.org:8080 \\[-0.25em]
      Accept: application/json
    };
    \draw[send] ($(server2)!0.55!(server2-ground)$) -- ($(client)!0.9!(client-ground)$)
    node[msg, pos=0.3] {
      200 OK HTTP/1.1 \\[-0.25em]
      Content-Type: application/json \\[0.1em]
      \{"ids": [14, 35, 28, 6, 22]\}
    };
  \end{tikzpicture}
  \caption{Hypermedia conversation between a broker and a registar}
  \label{fig:reg-conversation}
\end{figure}

To illustrate why, we notice that the example we used to present the
W3C HTTP Vocabulary in RDF is not realistic because, very often,
parameters in Web APIs are passed in the query string. Moreover the
link associated with the Location header is a Literal where we would
want to have it as a URI. We would like to express the conversation
shown in Figure~\ref{fig:reg-conversation}.

This conversation is composed of two interactions. The first one is
similar to the example of Figure~\ref{fig:reg-inter} but with the
difference that the number of requested identifiers takes the form of
a \verb=count= query parameter.  This difference is meaningful given
the wide adoption of this convention which is compatible with browser
form handling.  The second interaction consists in dereferencing the
link provided by the \verb=Location= header of the response of the
first interaction. The dereferenced link has a response that contains
a body content in the JSON format, which has its own structure which
is more complex than the plain string used in the request in Figure
\ref{fig:reg-inter}.  This is not possible to attach meaningful
semantics to this conversation by simply using the HTTP RDF
vocabulary.

Another aspect is that the HTTP RDF vocabulary is based on the
RFC~2616 which has been superseded by other RFC documents and in
particular by the RFC~7231. This topic of this updated specification
is dedicated to the semantics of HTTP~1.1. Those semantics have been
mostly preserved by newer versions of the protocols. Those upgrades
are mainly concerned by performance aspect in the transport layer
meaning either by optimizing TCP connections in the case of HTTP~2 of
replacing it with UDP in the case of HTTP~3.

Our proposals presented in next section aim at overcoming the
previously mentioned limitations, but in order to precisely evaluate
their usefulness we also devise a set of Competency Questions that we
would like to answer more easily with our proposed ontology.
Competency Questions reflect functional requirements representing
ontological commitments~\cite{Noy01ontologydevelopment,blomqvist2010,hitzler2016}.
\begin{competency}[Media type]
  \label{cq:msg-mt}
  What is the media-type associated with a message body?
\end{competency}
\begin{competency}[Interaction result]
  \label{cq:inter-result}
  What is the status code number of an interaction?
\end{competency}
\begin{competency}[Header values, e.g. location header]
  \label{cq:header-value}
  What is the target URI provided in the \verb=location= header of a
  response?
\end{competency}
Being able to answer to the previous questions enable to answer to the
following one:
\begin{competency}[Conversation result]
  \label{cq:conver-result}
  What is the status code of the combinaison of two interactions, the
  second query targeting the URI provided in the \verb=location=
  header of the first response?
\end{competency}
\begin{competency}[Content negotiation]
  \label{cq:cn}
  Does the media-type of a response body match one of those declared
  in the \verb=Accept= header of its corresponding request?
\end{competency}
We should be able to also query the body content when it is expressed in RDF:
\begin{competency}[Body content]
  \label{cq:body}
  What are the values associated with a given property $p$ inside the
  body of an HTTP message?
\end{competency}
We should be able to ask for parameters in the URI query string:
\begin{competency}[Query parameters]
  \label{cq:params}
  What is the value of a specific parameter, e.g. named "age", passed
  in the query string of the target resource?
\end{competency}




\section{Extending the HTTP ontology}
\label{sec:ontology}
We are using the $\mathcal{SROIQ^{(D)}}$ Description Logic, which is
 associated with OWL2 DL, to describe our
ontology. This formalisation is grounded in the former HTTP RDF
vocabulary by the EARL W3C Working Group \cite{koch_http_2017}. 
We use the RDF property/class terminology instead of the classical
role/concept terminology. We denote by $\top$ the class of all
individuals and by $\mathcal{D}$ the class of data literals meaning
the class of things that cannot be the subject of any of properties
and that can have only one interpretation. $\top$ and $\mathcal{D}$
classes are disjoints.

\subsection{Message}
\label{sec:message}

Communications over the HTTP protocol are based on request/response
messages exchanged between a client sending a request and a server
providing the response.  A message is defined by a collection of
headers which represent the metadata and a body which
contains the data payload of the message.  It is not required for a
message to have any header elements or a body content.
\[
Message \sqsubseteq \forall headers.Headers \sqcap \forall
body.Content \qquad \top\sqsubseteq{\leq_1}body.\top
\]
The property $body$ is functional. The details regarding the $Headers$
and $Content$ classes are provided in Section \ref{sec:headers} and
Section \ref{sec:body} respectively.  There are only two disjoint
kinds of messages which are either requests or responses.
\[
Message \equiv Request \sqcup Response \qquad Request \sqcap Response \sqsubseteq \bot
\]
We relate those two kinds of messages with the property $resp$ whose
domain is $Request$ and range is $Response$.
\[
\exists resp.\top \sqsubseteq Request \qquad   \top \sqsubseteq \forall resp.Response
\]
It is possible to have multiple responses for the same request with
some restrictions that are explained in Section \ref{sec:response}.

\subsection{Request}
\label{sec:request}

A request is a message which must have a method and an effective
URI. We define the class $Method$ as a superclass of all standardised
request methods using \emph{nominals}. We do not use equivalence
relationship because request methods can be extended.
\begin{gather*}
  \{ \textsc{get}, \textsc{head},
  \textsc{post}, \textsc{put}, \textsc{delete}, \textsc{connect},
  \textsc{options}, \textsc{trace}, \textsc{patch}\} \sqsubseteq Method \\
  Request \sqsubseteq Message \sqcap \exists mthd.Method \sqcap
  \exists uri.URI \\
  \top\sqsubseteq{\leq_1}mthd.Method \qquad \top\sqsubseteq{\leq_1}uri.URI
\end{gather*}
The properties $mthd$ and $uri$ are functional.  The value
associated with the property $uri$ is a URI following the syntax
described in RFC 3986 \cite[]{berners2005rfc} which provides the
following illustrative example:
\[
\underbrace{\texttt{http}}_{scheme}\texttt{://}\underbrace{\texttt{example.com:8042}}_{authority}
\underbrace{\texttt{/over/there}}_{path}\texttt{?}\underbrace{\texttt{name=ferret}}_{query}
\texttt{\#}\underbrace{\texttt{nose}}_{fragment}
\]
The property $uri$ abstracts the possible concrete syntaxes found in
the target URI which can take multiple forms
\cite[\href{https://tools.ietf.org/html/rfc7230\#section-5.3}{Section
    5.3}]{fieldingrfc}. Some of them require to combine the value of
the \verb+Host+ header and the protocol scheme (\verb=http= or
\verb=https=) to compute the absolute URI.  When the request target
has an absolute form, the target URI corresponds to the effective
request URI.  The various component parts of the URI associated with
the property $uri$ can be extracted from it with the properties
$scheme$, $authorithy$, $path$, $query$, $fragment$. Those properties
have literal values.  This choice differs from the W3C HTTP vocabulary
which uses the properties \verb=http:requestURI=,
\verb=http:absolutePath= and \verb=http:absoluteURI= to provide
different views of the target URI but represented as literals. The
property $uri$ corresponds to the same value as
\verb=http:absoluteURI= but lifted to an actual URI. This matters in
the context of RDF because only URIs can be the subject of
properties. This choice allows us to answer to CQ \ref{cq:params}
because we can associate to individuals of class URI a property
\verb=queryParams= to explicit the query parameters when they are
represented by the value associated with the property $query$, as
detailed in Section \ref{sec:query-params}.  The solution in the W3C
HTTP vocabulary is to represent the target URI as a string and this is
not enough to decompose the parameters.

\subsection{Query parameters}
\label{sec:query-params}

One important aspect when describing an HTTP request is to define
parameters, which can be passed in multiple ways but one basic way is
to use the query part of the URI meaning the characters between the \verb+?+
and \verb+#+. In the context of Web applications the format of this
part of the URI conforms to the
\verb=application/x-www-form-urlencoded=\footnote{\url{https://url.spec.whatwg.org/\#concept-urlencoded}}
media type which enables passing key-value pairs as arguments which
are denoted $k$ and $v$ in the following example.
\[
\underbrace{
  \overbrace{\texttt{age}}^{k}\texttt{=}\overbrace{\texttt{54}}^{v}
  \texttt{\&}\overbrace{\texttt{id}}^{k}\texttt{=}\overbrace{\texttt{XPZIJ4}}^{v}
}_{query}
\]
With the property $query$ we can access the string literal
corresponding to the encoded version.  We want to access those
key-value pairs semantically with proper properties and classes.  We
then define the property $queryParams$.
\begin{gather*}
  URI \sqsubseteq \forall queryParams.Parameter \qquad Parameter \equiv
  \exists name.\mathcal{D} \sqcap \exists value.\mathcal{D}
\end{gather*}
Our representation of URIs is depicted in
Figure~\ref{fig:our-vocab}.

\subsection{Response}
\label{sec:response}

A response is a message which must have a status instance accessible
via the property $sc$ which itself has a status code accessible via
the property $code$ whose value must be a 3 digit number. This number
is present in the status line of the concrete response.
\begin{gather*}
  Response \sqsubseteq Message \sqcap \exists sc.Status \qquad
  Status \sqsubseteq \exists code.\llbracket 000, 999\rrbracket \\
  \top\sqsubseteq{\leq_1}sc.\top \qquad
  \mathcal{D}\sqsubseteq{\leq_1}code.\mathcal{D}
\end{gather*}
The properties $sc$ and $code$ are functional. We use the compact
notation $\llbracket 000, 999\rrbracket$ to denote a non-negative
integer datatype with a restriction that its value is less or equal
than $999$.  This would take the following form in OWL 2 turtle syntax:
\begin{verbatim}
:threeDigit a rdfs:Datatype ;
    owl:equivalentClass [
        a rdfs:Datatype ;
        owl:onDatatype xsd:nonNegativeInteger ;
        owl:withRestrictions ([ xsd:maxInclusive 999 ])] .
\end{verbatim}
There exists a bijection between status instances and status codes
which means that the status code is characteristic of its instance.
For example we can define the status instance $Created$ which means
\emph{a new resource has successfully been created} by asserting that
this is the unique status instance with a status code of $201$.
\[
\{Created\} \equiv Status \sqcap \exists code.201
\]
Status codes can effectively be thought as the syntactic element that
denotes the meaning of the response status.  The meaning of the
response status is represented by the status instance. While each
status has specific meaning they can be classified in $Status$
subclasses that characterise the general result of the interaction.
\begin{gather*}
  Successful \equiv Status \sqcap \exists code.\llbracket 200, 299
  \rrbracket \quad
  ClientError \equiv Status \sqcap \exists code.\llbracket 400, 499
  \rrbracket \\
  Redirection \equiv Status \sqcap \exists code.\llbracket 300, 399
  \rrbracket \quad
  ServerError \equiv Status \sqcap \exists code.\llbracket 500, 599
  \rrbracket \\
  Informational \equiv Status \sqcap \exists code.\llbracket 100, 199
  \rrbracket
\end{gather*}
All the instances of those classes define \emph{final responses} with
the exception of the instance of the $Informational$ class which
defines \emph{interim responses} meaning temporary responses that will
eventually be followed by a \emph{final response}.
\cite[\href{https://tools.ietf.org/html/rfc7231\#section-6.2}{Section
    6.2}]{fielding2014hypertext}. This means that multiple responses
can be associated with one request but only one of them can be a
\emph{final response}.
\begin{gather*}
  \exists sc.Informational \sqsubseteq Interim \qquad
  Final \equiv Response \sqcap \lnot Interim
\end{gather*}

\begin{definition}[Interaction]
  An \emph{interaction} is an instance $resp(q, r)$ of the property $resp$ 
  such that $q$ is an instance of $Request$
  and $r$ is an instance of $Final$.
\end{definition}

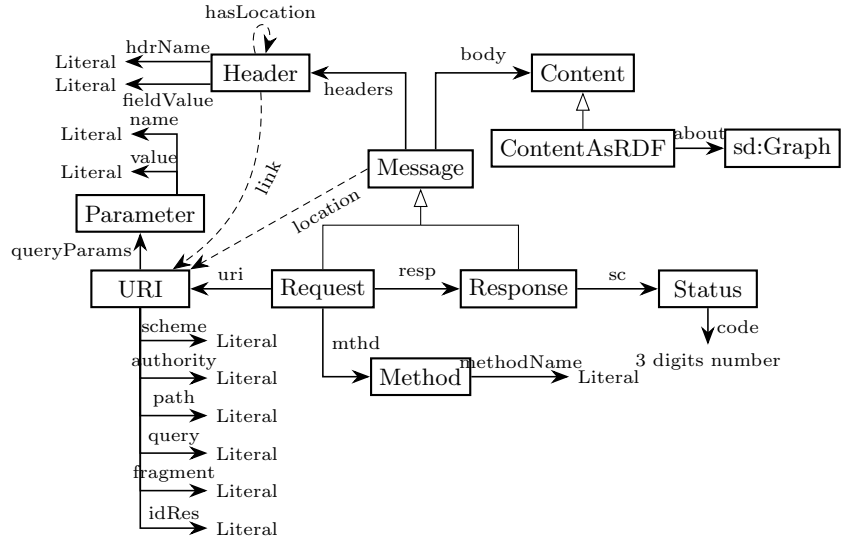
\begin{figure}
  \centering
  \begin{tikzpicture}
    \tikzset {
      class/.style={draw, solid, thick, font=\footnotesize,
        minimum height=1.5em, minimum width=4em, node distance=3.3em},
      lit/.style={font=\scriptsize, node contents=Literal, node distance=2.5em},
      prop/.style={arrows={-Stealth[length=7pt]}, semithick},
      lbl/.style={auto, font=\scriptsize}
    };
    \node[class] (Hdr) {Header};
    \node[class] (Msg) [below right=of Hdr] {Message};
    \node[coordinate] (MsgB) [below=1.5em of Msg] {};

    \node[class] (Cnt) [above right=of Msg] {Content};
    \node[class] (Rdf) [below=1.5em of Cnt] {ContentAsRDF};
    \draw[arrows = {-Stealth[inset=0em, fill=white,length=7pt]}] (Rdf) -- (Cnt);

    \node[class] (NG) [right=2em of Rdf] {sd:Graph};
    \draw[prop] (Rdf) to node[lbl] {about} (NG);

    \node[class] (Req) [below=of Msg, xshift=-4em] {Request};
    \node[class] (Mthd) [below=2em of Req, xshift=4em] {Method};
    \node[class] (Resp) [below=of Msg, xshift=4em] {Response};
    \node[class] (S) [right=of Resp] {Status};
    \node (sc) [font=\scriptsize, below=1.5em of S] {3 digits number};

    \draw[prop] (Resp) to node[lbl] {sc} (S);
    \draw[prop] (S) -- node[lbl] {code} (sc);

    \node (HdrT) [coordinate, above=1.5mm of Hdr.west] {};
    \node (HdrB) [coordinate, below=1.5mm of Hdr.west] {};
    \node (pN) [lit, left=of HdrT, xshift=-1em];
    \node (pV) [lit, left=of HdrB, xshift=-1em];
    \draw[prop] (HdrT) to node[lbl, above] {hdrName} (pN);
    \draw[prop] (HdrB) to node[lbl, below] {fieldValue} (pV);

    \node (mN) [lit, right=of Mthd, xshift=1.5em];
    \draw[prop] (Mthd) to node[lbl] {methodName} (mN);

    \node (uri) [class, left=of Req] {URI};
    \draw[prop] (Req) to node[lbl, above] {uri} (uri);

    \node (scheme) [lit, below right=1em of uri];
    \node (auth) [lit, below=0.3em of scheme];
    \node (path) [lit, below=0.3em of auth];
    \node (query) [lit, below=0.3em of path];
    \node (fragment) [lit, below=0.3em of query];
    \node (idRes) [lit, below=0.3em of fragment];
    \draw[prop] (uri) |- node[lbl, near end] {scheme} (scheme);
    \draw[prop] (uri) |- node[lbl, near end] {authority} (auth);
    \draw[prop] (uri) |- node[lbl, near end] {path} (path);
    \draw[prop] (uri) |- node[lbl, near end] {query} (query);
    \draw[prop] (uri) |- node[lbl, near end] {fragment} (fragment);
    \draw[prop] (uri) |- node[lbl, near end] {idRes} (idRes);

    \node (param) [class, above=1.5em of uri] {Parameter};
    \draw[prop] (uri) to node[lbl] {queryParams} (param);

    \node (paramT) [coordinate, right=1.5em of param.north] {};
    \node (PValue) [lit, above=0.3em of param, xshift=-2em];
    \node (PName) [lit, above=0.3em of PValue];
    \draw[prop] (paramT) |- node[lbl, near end, above] {name} (PName);
    \draw[prop] (paramT) |- node[lbl, near end, above] {value} (PValue);

    \draw[arrows={-Stealth[length=7pt]}, densely dashed] (Hdr)
    to[loop above] node[lbl] {hasLocation} (Hdr);
    \draw[arrows={-Stealth[length=7pt]}, densely dashed] (Hdr)
    to[bend left, sloped] node[lbl] {link} (uri);
    \draw[arrows={-Stealth[length=7pt]}, densely dashed] (Msg.west)
    to node[lbl, sloped] {location} (uri.north east);

    \draw (Req) |- (MsgB);
    \draw (Resp) |- (MsgB);
    \draw[arrows = {-Stealth[inset=0em, fill=white,length=7pt]}] (MsgB) -- (Msg);

    \draw[prop] (Msg) [xshift=-3em] |- node[near end, lbl] {headers} (Hdr);
    \draw[prop] (Msg) [xshift=3em] |- node[lbl, near end] {body} (Cnt);
    \draw[prop] (Req) to node[lbl] {resp} (Resp);
    \draw[prop] (Req) [yshift=-1em] |- node[lbl,near start] {mthd} (Mthd);
  \end{tikzpicture}
  \caption{Main elements of our ontology for HTTP.}
  \label{fig:our-vocab}
\end{figure}

In Figure~\ref{fig:our-vocab}  we do not represent the sub-classes of 
Status, as we do not represent those of StatusCode in Figure~\ref{fig:rdf-vocab}.

\subsection{Header fields}
\label{sec:headers}

A Message can contain multiple headers. A header is composed of a
field name and a field value which are both literals.
\begin{gather*}
  Header \sqsubseteq \exists \mathit{hdrName}.\mathcal{D} \sqcap \exists \mathit{fieldValue}.\mathcal{D}
\end{gather*}
Such semantic view is still not that descriptive because of the usage
of literals which are just strings. To refine the semantic of standard
headers 
such as \verb=Location= and \verb=Content-Type=, we can use
a property chain in order to directly refer to their values with properties
reflecting their names. This is motivated by our Competency Questions
CQ \ref{cq:msg-mt} and CQ \ref{cq:header-value}.  It is common to have a location header such as in the
following example:
\[
\verb+Location: <http://example.org/new/resource>+
\]
We want to access simply to the associated link.  We can then refine
the corresponding header instances by stating that if its header field
name is \verb+"Location"+ then it has a $link$ property.  The value of
that property corresponds to the URI literal contained in the property
$fieldValue$, but lifted to an URI. This enables the definition of
properties on that value.
\[
Header \sqcap \exists \textit{hdrName}.\{\texttt{{\small"Location"}}\}
\sqsubseteq \exists link.\top \\
\]
In order to describe the associative nature of message headers we can
use property chain axioms to associate request to the lifted semantic
value associated with standards headers.  For example for the
\verb=Location= header we can define the property $location$.  To
achieve that we introduce a reflexive property $hasLocation$ for only
those instances that have a \verb=Location= field name.
\begin{gather*}
  \exists \textit{hdrName}.\{\texttt{{\small"Location"}}\} \equiv \exists hasLocation.Self \\
  headers \circ hasLocation \circ link \sqsubseteq location
\end{gather*}
This mechanism is illustrated in Figure~\ref{fig:our-vocab} for the 
\verb=Location= header. We propose to do that for
all predefined headers, and in particular the \verb=Content-Type= header. 

The W3C HTTP vocabulary approach is different in that regard because
it tries to augment a literal view of headers with the property
\verb=http:headerElements= whose values can be decomposed with
\verb=http:elementName=, \verb=http:elementValue= and
\verb=http:params= in a generic way.  The representation of
heterogeneous associative data structures like message headers by a
conjonction of well characterized property constraints like the
\verb=Location= header is more precise despite requiring a wider TBox.

\subsection{Body Content}
\label{sec:body}

While the header fields are the representation metadata, the payload
body is the representation data. As defined in Section
\ref{sec:message}, this data is accessible from a message via the
property $body$ which value is always of class $Content$.  The W3C
HTTP vocabulary delegates the representation of the body values to an
external Content Vocabulary \cite{koch2011representing} which takes
into account the fact that a resource can be associated with multiple
representations in various formats. Unfortunately the HTTP W3C
vocabulary restricts the range of the property $body$ to
$ContentAsBase64$ which does not seem to match the semantics of HTTP
where the actual format of the body is advertised by the
\verb=Content-Type= and \verb=Content-Encoding= headers. The first one
is mandatory when having a body and its value is a media type
$\mathcal{M}$.  Those headers allow the recipient of a message to know
how to interpret the body content. For example when a
\verb=Content-Type= of \verb=text/plain= and no
\verb=Content-Encoding= is present the $body$ value should directly be
an instance of the $ContentAsText$ class.
\[
\exists body.Content \sqsubseteq
\exists \texttt{\footnotesize{content-type}}.\mathcal{M}
\]
In CQ \ref{cq:inter-result} we want to access the links provided in a
message. Some of those links are provided in the header field values
but others are provided in the body. This is true when the
\verb=Content-Type= corresponds to an hypermedia format. Since RDF
provides a native way to represent those links which are plain URIs,
our proposal is to require the body content to be available as
RDF. To do that, we must clearly encapsulate the content graph in order 
to distinguish it from the graph of the interaction. With that restriction 
we can have both the graph of the body
and the graph of its message container in a single formalism.  We
implement that requirement by introducing the $ContentAsRDF$ class
as a sub-class of Content. Property $about$ serves as the
bridge between the message and the content's graph, and this latter is 
a named graph as defined in the SPARQL 1.1 Service 
Description\footnote{\url{https://www.w3.org/TR/2013/REC-sparql11-service-description-20130321/}}.
\begin{gather*}
ContentAsRDF \sqsubseteq Content \sqcap \exists about.\top \quad
\top\sqsubseteq \forall about.Graph \quad
\top\sqsubseteq{\leq_1}about.\top
\end{gather*}
The property $about$ is functional. 
For example a body message describing Resource \verb=:foo= can be 
represented with the following TriG syntax:
\begin{verbatim}
:B {
    :foo :ids (1 2 3) ;
         :date "2003-02-10"^^xsd:date .
}
\end{verbatim}
Then it can be associated with the body content in the following way:
\begin{verbatim}
:m a http:Message ;
   http:body :b .
:b a cnt:ContentAsRDF ;
   cnt:about :B .
:B a sd:Graph .
\end{verbatim}
For having the body payload in RDF, the basic option is to require
representations to be in RDF serialisation formats like RDF/XML,
JSON-LD and Turtle. A more advanced option is to adopt the notion of
\emph{RDF presentation} \cite{lefranccois2018rdf}, which provides a
way to \emph{lift} a non-RDF format into an RDF graph and a reverse
way to \emph{lower} an RDF graph representation into the same non-RDF
format.


\section{Evaluation}
\label{sec:eval}
\begin{figure}
  \centering
  \begin{tikzpicture}
    \tikzset{
      I/.style={draw, rounded corners},
      C/.style={draw, font=\scshape, thick},
      L/.style={font=\itshape\ttfamily\scriptsize, fill=black!20},
      E/.style={draw, circle},
      every edge quotes/.style={sloped,above,font=\scriptsize}
    };
    \node[I] (q1) at (-0.5, 0) {$q1$};
    \node[I] (r1) at (1.5, 0) {$r1$};
    \node[I] (post) at (1.5, 1) {\textsc{post}};
    \node[I] (reg) at (-3, -0.5) {$uri1$};
    \node[L] (regRes) at (-1.5, 1) {\texttt{\scriptsize{\ldots/reg?count=5}}};
    \node[L] (res2) at (3, -1.8) {\texttt{\scriptsize{\ldots/reg/x8344}}};
    \node[E] (params) at (-0.5, -1.7) {};
    \node[L] (param) at (1, -0.7) {count};
    \node (count) at (1.2, -1.8) {$5$};
    \node[I] (created) at (3.5, 0.9) {$Created$};
    \node[I] (q2) at (4, 0) {$q2$};
    \node[I] (uri2) at (5.5, -1.2) {$uri2$};
    \node[I] (get) at (6, 1) {\textsc{get}};
    \node[I] (r2) at (6, 0) {$r2$};
    \node[I] (ok) at (7.5, 0.7) {$Ok$};
    \node[E] (body) at (7.5, -0.4) {};

    \node[I] (cnt) at (7.5, -1.5) {\texttt{\scriptsize{/reg/x8344}}};
    \node (list) at (10, -1.5) {$[14, 35, ...]$};
    \node[draw, rounded corners, fit=(cnt) (list)] (graph) {};

    \draw[->] (uri2) edge["$idRes$"] (res2);
    \draw[->] (reg) edge["$idRes$"] (regRes);
    \draw[->] (q1) edge["$resp$"] (r1);
    \draw[->] (q1) edge["$uri$"] (reg);
    \draw[->] (q1) edge["$mthd$"] (post);
    \draw[->] (reg) edge["$queryParams$"] (params);
    \draw[->] (params) edge["$name$"] (param);
    \draw[->] (params) edge["$value$"] (count);
    \draw[->] (r1) edge["$sc$"] (created);
    \draw[->] (r1) edge["$location$"] (uri2);
    \draw[->] (q2) edge["$uri$"] (uri2);
    \draw[->] (q2) edge["$mthd$"] (get);
    \draw[->] (q2) edge["$resp$"] (r2);
    \draw[->] (r2) edge["$sc$"] (ok);
    \draw[->] (r2) edge["$body$"] (body);
    \draw[->] (body) edge["$about$"] (graph);
    \draw[->] (cnt) edge["$ids$"] (list);
  \end{tikzpicture}
  \caption{Graph representation of registar conversation}
  \label{fig:reg-conv}
\end{figure}
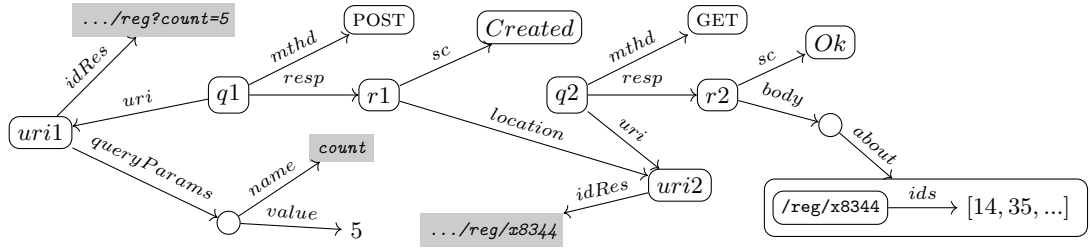

One important step to enable evaluation is to populate the ontology with 
representative individuals. We can now represent the two interactions 
described in Section \ref{sec:pbstate} and illustrated in Figure \ref{fig:reg-conversation}.
A visual representation of the associate RDF graph is provided in Figure
\ref{fig:reg-conv}.

The question of ontology evaluation has received a lot of attention 
 and can be divided in several related categories~\cite{Tartir2010}, 
 logical, structural, and functional. The \textit{logical} category groups 
 quality dimensions that can be performed using a reasoner, e.g. the
  satisfiability. Our implementation in OWL 2 DL allows us to verify
  inference capabilities with the HermiT reasoner, in particular capability
   of detecting logical inconsistencies through \textit{error provocations} 
   during the population of the ontology with some representative sets of individuals. 
 The \textit{structural} category is composed of 
 context-free dimensions which can be measured by quantitative metrics, 
 for instance using 
 OntoMetrics\footnote{\url{https://ontometrics.informatik.uni-rostock.de/ontologymetrics}}, 
 but also the popularity
 or the coupling degree with other linked data resources. 
 To us, some important dimensions in this category are the following ones:
 \begin{itemize}
\item flexibility: is the ontology easily adaptable to multiple uses? We address 
this question with our proposals for headers, body content and query parameters
which enhance the reusability potential of the ontology with respect to 
the original W3C HTTP RDF Vocabulary. 
\item transparency: is the ontology easily analysable? We address this question 
by our formalisation in Description Logic, implemented with OWL2 DL,  
explicating our choices and motivations.
\item cognitive ergonomics: is the ontology easily understandable 
and exploitable by users? We address this question by writing this article,
documenting the ontology, and publicly publishing it.
\item compliance to expertise: is the ontology compliant with the 
knowledge it represents. In our case, we better represent
the semantics of RFCs by introducing classes and properties in place of
just string literals, for instance for URIs. This dimension is  also
related to functional properties which, for an application ontology 
such as the one presented in this paper, are of course of outmost importance.
\end{itemize} 

The \textit{functional} category groups ontology quality dimensions which
address intended uses and functions in contexts.
We address this by writing SPARQL queries on our ontology
in order to answer to the Competency Questions we expressed in
Section~\ref{sec:pbstate}.

\begin{description}
\item [CQ \ref{cq:msg-mt}] What is the media-type associated with a
  message body?
\begin{verbatim}
SELECT ?m ?mt
WHERE {
  ?m a http:Message .
  ?m http:content-type ?mt .
}
\end{verbatim}

\item[CQ \ref{cq:inter-result}] What is the status code number of an interaction?
\begin{verbatim}
SELECT ?status
WHERE {
  ?q http:resp ?r .
   ?r http:sc/http:code ?status .
}
\end{verbatim}

\item[CQ \ref{cq:header-value}] What is the URI link provided in the
  \verb=location= header of a response?
\begin{verbatim}
SELECT ?next
WHERE {
  ?q0 http:resp/http:location ?next .
}
\end{verbatim}

\item[CQ \ref{cq:conver-result}] What is the status code of the
  combinaison of two interactions, the second query targeting the URI
  provided in the \verb=location= header of the first response?
\begin{verbatim}
SELECT ?status
WHERE {
  ?q0 http:resp/http:location ?next .
  ?q1 http:uri ?next .
  ?q1 http:resp/http:sc ?status .
}
\end{verbatim}

\item[CQ \ref{cq:cn}] Does the media-type of a response body match one of
  those declared in the \verb=Accept= header of its corresponding
  request?
\begin{verbatim}
ASK {
  ?q http:resp ?r .
  ?q http:accept/http:media-type ?mt1 .
  ?r http:content-type ?mt2 .
  FILTER (CONTAINS(STR(?mt1), STR(?mt2))
          || CONTAINS(STR(?mt2), STR(?mt1)))
}
\end{verbatim}
Here we rely on string comparison which is a simplistic form of content
negotiation because in practice media types have derivatives with
syntactic variations.  A semantic description of media types would
solve that limitation.

\item[CQ \ref{cq:body}] What are the values associated with a given
  property $p$, for instance \verb=ex:ids=,  inside the body of an HTTP message?
\begin{verbatim}
SELECT ?ids
WHERE {
  ?m http:body/cnt:about ?G .
  GRAPH ?G { ?x ex:ids/rdf:rest*/rdf:first ?ids } .
}
\end{verbatim}
Here Property \verb=ex:ids= associates the root of the content representation
with an RDF list of identifiers. Each element present in that list is
extracted by using the \verb=rdf:rest*/rdf:first= pattern.

\item[CQ \ref{cq:params}] What is the value of a specific parameter,
  e.g. named "age", passed in the query string of the target resource?
\begin{verbatim}
SELECT ?age
WHERE {
  ?q http:uri/http:queryParams ?p .
  ?p http:name "age" .
  ?p http:value ?age
}
\end{verbatim}

\end{description}


\section{Related Work}
\label{sec:related}
As explained in Section~\ref{sec:pbstate} our aim is to validate a Web API
with respect to a specification, in order to support the work of developers,
producers and consumers of Web APIs. Web developers know well HTTP and
are less interested in more abstracted representations: for explaining
them their potential mistakes, we think that it is better for the validator
to manipulate HTTP items. This is why we choose to rely on an ontology of HTTP.
It allows us to represent a Web API as a \textit{conversation} graph, composed of
linked interactions, an \textit{interaction} being an instance of Property $resp$
relating an instance of class $Request$ to an instance of class $Response$.
The HTTP ontology provides the framework for building the conversation graph
from the Web API code.
It is also a guide for the automatic validation that we intend to perform on the
conversation graph, through the verification of the constraints expressed in the
specification.

In this related works section we consider the Web API specification problem which
motivates our work, from several points of view, reflected by the past and current
uses of the various proposals. Historically, at the beginning were
the so-called Web services, defined with SOAP, WSDL and UDDI~\cite{WSDL2005},
with several proposals aimed at specifying semantic Web services,
such as OWL-S~\cite{owl-s2005} or SAWSDL~\cite{SAWSDL2007}.
These Web services follow a remote procedure call (RPC) form of inter-process
communication popularised by the rise of object oriented programming, ignoring the
principle of hypermedia at the heart of the Web, and defining and exposing their own
arbitrary sets of operations rather than the defined HTTP methods.
They have been widely studied and developed, and big enterprise
information systems still use it (e.g. Amazon).
From a developer point of view WSDL specifications
really provide a good support for using such kind of services,
but no attempt to automate their
use by programs has definitely proved successful.
Yet WSDL provides a machine-readable description of how the service can be called,
what parameters it expects, and what data structures it returns~\cite{WSDL2005}.
It describes services as collections of network endpoints (ports), which are
associations of a URL with a binding describing a concrete protocol and
message format specifications, namely the supported operations with
descriptions of the data being exchanged. SAWSDL~\cite{SAWSDL2007} is a set
of extension attributes for
defining how to add semantic annotations to the various parts of a WSDL document.
It allows designers to relate WSDL and XML Schema specifications to ontologies.

The subsequent history of Web services is dominated by the so-called REST services or
RESTful services~\cite{richardson2013restful}, or hypermedia
APIs~\cite{VerborghAHRMSG17}, together with the emergence of Linked Data and
the growing popularity of JSON and JSON-LD.
Before considering these other kinds of Web services, it is important to notice the
universality of the requirement of a way to provide information on a
service general functionalities as well as on the form and meaning of its inputs and outputs.
This is the aim of SAWSDL for SOAP-based services, or OWL-S~\cite{owl-s2005}
intended for any kind of service. Interestingly, this one can be
compared to the recent proposal of another universal description of
functions~\cite{functionOntology2016}, which is also motivated to complement
Web services specifications that are very coupled with the technology
stack\footnote{\url{https://w3id.org/function/spec}}, allowing to declare and describe
any specific problem, function, or algorithm, whether executed using Web APIs or not.
Both OWL-S and the Function Ontology address the need to automatic discovery,
invocation, composition and interoperation of services.
OWL-S is an old W3C Member Submission (2004)
while the Function Ontology is at a stage of first non official draft,
intended for applications built on top of the Semantic Web.
Both of them include a binding mechanism to relate the upper level functional
descriptions to more concrete Web APIs descriptions, these latter being expressed
using WSDL, or the HTTP ontology, or Hydra~\cite{lanthaler2013hydra}.

Hydra's purpose is to describe and use the so-called Web APIs that are based on the
REST architectural style and are simpler to deploy and interact with than SOAP-based
services~\cite{upadhyaya2011migration,richardson2013restful}.
Those Web APIs are now the most developed
and used Web services. OpenAPI~\cite{initiativeopenapi} is a widely used initiative
to associate to Web APIs a machine-readable documentation, that can be compiled
into a Web page, and thus findable with a Web browser using standard Web search
engines. With such a syntax-level description, OpenAPI offers what WSDL enables
for SOAP Web services, or WADL for REST Web services (JSON replacing XML).
In contrast, Hydra offers a semantic-level description and aims to go further in
simplifying the development of truly RESTful services by leveraging the power of
Linked Data~\cite{lanthaler2013hydra},
which is mostly ignored by the OpenAPI initiative.
To this end, Hydra consists in a lightweight RDFS vocabulary to both describe Web APIs and
to augment Linked Data with hypermedia controls (allowing to specify which IRIs
in an RDF graph are intended to be dereferenced or not).
Notice that W3C also supports the Linked Data Platform (LDP)
standard~\cite{LDP2015} whose aim is to allow Linked Data (LD) providers exposing
LD in a RESTful manner, including an interaction model to interact (read-write) with them.
This is in line with Hydra's intention of describing RESTful Web APIs that can consume LD.
Hydra is the most similar proposal to the HTTP ontology.
But it is different though, as it introduces more abstract classes and properties to represent
HTTP and IRI components, together with purely RDF-based notions which requires
Web API developers to master (JSON-LD serialised) RDF.

Hydra is used in several applications, for instance for automating the discovery
and consumption of Web services by software agents via SPARQL
micro-services~\cite{MichelFCG19}, which allow programs to query any
Web API using SPARQL.
The Web API is wrapped by such a micro-service, which uses an
internally stored description of the Web API
and the incoming SPARQL query to (i) query the Web API, (ii) transform
its output into an RDF graph and (iii) perform the query on this dynamically built
graph. The Web API functional description is based on Hydra and schema.org,
plus SHACL~\cite{MichelFCG19} for guiding the dynamic graph building.

RESTdesc~\cite{verborgh2012functional,VerborghAHRMSG17}
is also a semantic description format for hypermedia APIs, which relies on the HTTP
ontology rather than Hydra, and on Notation3 Logic~\cite{BernersLee2007N3LogicAL}
rather than RDFS or OWL.
It is devised for automatic discovery and composition of hypermedia APIs.
For its authors, the intended benefit of using Linked Data is to
provide the ground to enable intelligent agents to navigate APIs and
perform choices at runtime, like humans do on Website. While this goal is
highly interesting, it is far from what is done in practice in current
enterprise information systems
where the service composition is manually programmed from a fixed set
of endpoints. In our experience this careful manual composition process
is already making things
difficult to manage and maintain when services evolve,
to not want to rely on automatic compositions as proposed
in RESTdesc to fulfil our requirements.
In order to support the daily work of producers and consumers of Web APIs,
we basically want to validate a Web API with respect to a specification,
and explain the possibly occurring errors to the developer. To this end
we focus on representing the common knowledge shared by
all these developers: HTTP interactions.


\section{Conclusion}
\label{sec:conclusion}
We present in this article an ontological description of HTTP interactions,
based on a deep study of
the HTTP RFC 7231 specification. 
We started from the HTTP RDF Vocabulary proposed as a W3C Note, 
used and reused by several works. Besides that, it relies on a limited
interpretation (more attached to the HTTP syntax than semantics) of  
RFCs, it does not allow us to formulate simply the queries which 
correspond to the Competency Questions reflecting our requirements.
We conducted a formal analysis, using both Description Logics and 
Protégé OWL 2 DL, to introduce our proposed ontology, which 
answers to our needs as shown by our evaluation efforts.
Regarding the related works, to the best of our 
knowledge only the W3C HTTP RDF Vocabulary is directly
related to our aim of semantically representing the HTTP protocol,
while the more general aim of describing web services is addressed 
in many works. This more general problem is not directly usable 
to validate a Web API with respect to a HTTP specification, 
for simply explaining to developers their potential errors. 

The HTTP RFC 7231 specification being quite large,
our effort to formalise it still incomplete and multiple aspect
remains to be studied. Moreover we observed in the case of query
parameters that common practices are in fact ad-hoc extensions of the
URI specification, but they still need to be taken into account because of
their massive adoption.
We are aware of several limitations of our approach. For instance when
representing an HTTP conversation that involves multiple interactions on
the same resource, we still are unable to describe the evolution of those
representations. Neither do we define how to represent the time 
dependency relation between messages. 
Another limitation comes from the Open World Assumption (OWA) 
which means that it is not possible to check for the absence of things, 
which is useful in a context of validation. For example we could want 
to check that responses associated with the \verb=HEAD= method
do not have a body. Combining OWL and SHACL could be the solution.
Indeed, our immediate future work is the exploitation of
that ontology by instrumenting HTTP interactions either on the client
or on the server side. The logical inference performed by OWL 2 reasoner
will check the conformance of those interactions towards the protocol
specification. For example it could check for every message that
\verb=Content-Type= are properly declared in the presence of a
body. To achieve this, we are working to implement a program that
converts HTTP messages into RDF graphs following our ontology.


\bibliographystyle{splncs04}
\bibliography{biblio}

\begin{thebibliography}{10}
\providecommand{\url}[1]{\texttt{#1}}
\providecommand{\urlprefix}{URL }
\providecommand{\doi}[1]{https://doi.org/#1}

\bibitem{BernersLee2007N3LogicAL}
Berners-Lee, T., Connolly, D., Kagal, L., Scharf, Y., Hendler, J.A.: N3logic: A
  logical framework for the world wide web. Theory Pract. Log. Program.
  \textbf{8},  249--269 (2007)

\bibitem{berners2005rfc}
Berners-Lee, T., Fielding, R., Masinter, L.: Uniform resource identifier (uri):
  Generic syntax (2005)

\bibitem{blomqvist2010}
Blomqvist, E., Presutti, V., Daga, E., Gangemi, A.: Experimenting with extreme
  design. In: Cimiano, P., Pinto, H.S. (eds.) Knowledge Engineering and
  Management by the Masses. pp. 120--134. Springer Berlin Heidelberg, Berlin,
  Heidelberg (2010)

\bibitem{functionOntology2016}
De~Meester, B., Dimou, A., Verborgh, R., Mannens, E.: {An Ontology to
  Semantically Declare and Describe Functions}. In: The Semantic Web. pp.
  46--49. Springer International Publishing, Cham (2016),
  \url{https://w3id.org/function/spec}

\bibitem{fieldingrfc}
Fielding, R., Reschke, J.: Hypertext transfer protocol (http/1.1): Message
  syntax and routing (2014)

\bibitem{fielding2014hypertext}
Fielding, R., Reschke, J.: Hypertext transfer protocol (http/1.1): Semantics
  and content (2014)

\bibitem{gfielding2002principled}
Fielding, R.T., Taylor, R.N.: Principled design of the modern web architecture.
  ACM Transactions on Internet Technology (TOIT)  \textbf{2}(2),  115--150
  (2002)

\bibitem{Guarino1998}
Guarino, N.: Formal ontology and information systems. In: Proceedings of Formal
  Ontology in Information Systems. pp. 3--15. IOS Press (1998)

\bibitem{hitzler2016}
Hitzler, P., Gangemi, A., Janowicz, K., Krisnadhi, A., Presutti, V. (eds.):
  Ontology Engineering with Ontology Design Patterns - Foundations and
  Applications, vol.~25. IOS Press (2016)

\bibitem{initiativeopenapi}
Initiative, O.: {The OpenAPI Specification} (2018)

\bibitem{koch2011representing}
Koch, J., Velasco, C., Ackermann, P.: Representing content in rdf 1.0 (2017)

\bibitem{koch_http_2017}
Koch, J., Velasco, C., Ackermann, P.: {HTTP} {Vocabulary} in {RDF} 1.0 (Feb
  2017), \url{https://www.w3.org/TR/HTTP-in-RDF/}, {W3C} Working Group

\bibitem{SAWSDL2007}
Kopecký, J., Vitvar, T., Bournez, C., Farrell, J.: {SAWSDL}: semantic
  annotations for {WSDL} and {XML} schema. Internet Computing, IEEE
  \textbf{11},  60--67 (12 2007). \doi{10.1109/MIC.2007.134}

\bibitem{lanthaler2013hydra}
Lanthaler, M., G{\"u}tl, C.: {Hydra: A Vocabulary for Hypermedia-Driven Web
  API}s. LDOW  \textbf{996} (2013)

\bibitem{lefranccois2018rdf}
Lefran{\c{c}}ois, M.: {RDF} presentation and correct content conveyance for
  legacy services and the web of things. In: Proceedings of the 8th
  International Conference on the Internet of Things. p.~43. ACM (2018)

\bibitem{lirzin2020}
Lirzin, M., Markhoff, B.: {Vers une ontologie des interactions HTTP}. In: {31es
  Journ{\'e}es francophones d'Ing{\'e}nierie des Connaissances}. {S{\'e}bastien
  Ferr{\'e}}, Angers, France (Jun 2020),
  \url{https://hal.archives-ouvertes.fr/hal-02888065}

\bibitem{owl-s2005}
Martin, D., et., a.: {Bringing Semantics to Web Services: The OWL-S Approach}.
  In: Semantic Web Services and Web Process Composition. pp. 26--42. Springer
  Berlin Heidelberg, Berlin, Heidelberg (2005)

\bibitem{MichelFCG19}
Michel, F., Faron{-}Zucker, C., Corby, O., Gandon, F.: Enabling automatic
  discovery and querying of web {API}s at web scale using linked data
  standards. In: Companion of The 2019 World Wide Web Conference, {WWW} 2019,
  San Francisco, CA, USA, May 13-17, 2019. pp. 883--892 (2019).
  \doi{10.1145/3308560.3317073}, \url{https://doi.org/10.1145/3308560.3317073}

\bibitem{Noy01ontologydevelopment}
Noy, N., Mcguinness, D.: Ontology development 101: A guide to creating your
  first ontology. Knowledge Systems Laboratory  \textbf{32} (01 2001)

\bibitem{richardson2013restful}
Richardson, L., Amundsen, M., Ruby, S.: {REST}ful {W}eb {API}s. O’Reilly
  Media (2013)

\bibitem{LDP2015}
Speicher, S., Arwe, J., Malhotra, A.: Linked data platform 1.0, w3c
  recommendation. Tech. rep., W3C (2015)

\bibitem{Tartir2010}
Tartir, S., Arpinar, I.B., Sheth, A.P.: Ontological evaluation and validation.
  In: Theory and applications of ontology: Computer applications, pp. 115--130.
  Springer (2010)

\bibitem{upadhyaya2011migration}
Upadhyaya, B., Zou, Y., Xiao, H., Ng, J., Lau, A.: Migration of {SOAP}-based
  services to {REST}ful services. In: 2011 13th IEEE International Symposium on
  Web Systems Evolution (WSE). pp. 105--114. IEEE (2011)

\bibitem{VerborghAHRMSG17}
Verborgh, R., Arndt, D., Hoecke, S.V., Roo, J.D., Mels, G., Steiner, T.,
  Gabarr{\'{o}}, J.: The pragmatic proof: Hypermedia {API} composition and
  execution. Theory Pract. Log. Program.  \textbf{17}(1),  1--48 (2017).
  \doi{10.1017/S1471068416000016},
  \url{https://doi.org/10.1017/S1471068416000016}

\bibitem{verborgh2012functional}
Verborgh, R., Steiner, T., Van~Deursen, D., Coppens, S., Vall{\'e}s, J.G.,
  Van~de Walle, R.: Functional descriptions as the bridge between hypermedia
  {API}s and the semantic web. In: Proceedings of the third international
  workshop on {REST}ful design. pp. 33--40. ACM (2012)

\bibitem{WSDL2005}
Weerawarana, S., Curbera, F., Leymann, F., Storey, T., Ferguson, D.F.: {Web
  Services Platform Architecture: SOAP, WSDL, WS-Policy, WS-Addressing,
  WS-BPEL, WS-Reliable Messaging and More}. Prentice Hall PTR, USA (2005)

\end{thebibliography}

\appendix

\section{Appendix: HTTP Interaction Ontology in Turtle}
\label{sec:appendix}
\begin{verbatim}
@prefix : <http://w3id.org/http#> .
@prefix mthd: <http://w3id.org/http/mthd#> .
@prefix sc: <http://w3id.org/http/sc#> .
@prefix hds: <http://w3id.org/http/headers#> .
@prefix cnt: <http://w3id.org/http/content#> .

@prefix owl: <http://www.w3.org/2002/07/owl#> .
@prefix rdfs: <http://www.w3.org/2000/01/rdf-schema#> .
@prefix rdf: <http://www.w3.org/1999/02/22-rdf-syntax-ns#> .
@prefix sd: <http://www.w3.org/ns/sparql-service-description#> .
@prefix xsd: <http://www.w3.org/2001/XMLSchema#> .

: a owl:Ontology ;
    rdfs:label "HTTP Ontology"@en ;
    rdfs:comment "A namespace for describing HTTP interactions"@en .

## --------- ##
## Messages. ##
## --------- ##

:Message a owl:Class, owl:AllDisjointClasses ;
    rdfs:label "Message"@en ;
    rdfs:comment "An HTTP message."@en ;
    rdfs:isDefinedBy <http://tools.ietf.org/rfc/rfc7231> ;
    owl:members (:Request :Response) ;
    rdfs:subClassOf [
        owl:intersectionOf ([
                a owl:Restriction ;
                owl:onProperty :body ;
                owl:allValuesFrom cnt:Content ;
                ] [
                a owl:Restriction ;
                owl:onProperty :hdr ;
                owl:allValuesFrom :Header;
                ])] .

:Request a owl:Class ;
    rdfs:label "Request"@en ;
    rdfs:comment "An HTTP request."@en ;
    rdfs:subClassOf [
        owl:intersectionOf (:Message [
                a owl:Restriction ;
                owl:onProperty :mthd ;
                owl:someValuesFrom :Method ;
                ] [
                a owl:Restriction ;
                owl:onProperty :uri ;
                owl:someValuesFrom :URI;
                ])] ;
    rdfs:isDefinedBy <http://tools.ietf.org/rfc/rfc7231> .

:Response a owl:Class, owl:AllDisjointClasses ;
    rdfs:label "Response"@en ;
    rdfs:comment "An HTTP response."@en ;
    owl:members (:InterimResponse :FinalResponse) ;
    rdfs:subClassOf [
        owl:intersectionOf (:Message [
                a owl:Restriction ;
                owl:onProperty :sc ;
                owl:someValuesFrom :StatusCode ;
                ])] .

:InterimResponse a owl:Class ;
    rdfs:label "Interim"@en ;
    rdfs:subClassOf :Response, [
        a owl:Restriction ;
        owl:onProperty :sc ;
        owl:someValuesFrom sc:Informational ;
        ] ;
    rdfs:comment "An interim response."@en .

:FinalResponse a owl:Class ;
    rdfs:label "Final"@en ;
    rdfs:subClassOf :Response, [ owl:complementOf :InterimResponse ] ;
    rdfs:comment "A final response."@en .

:resp a owl:ObjectProperty ;
    rdfs:label "response"@en ;
    rdfs:comment "The HTTP response sent in answer to an HTTP request."@en ;
    rdfs:domain :Request ;
    rdfs:range :Response .

## ------- ##
## Method. ##
## ------- ##

:Method a owl:Class ;
    rdfs:label "Method"@en ;
    rdfs:comment "The HTTP method used for the request."@en ;
    owl:equivalentClass [
        a owl:Restriction ;
        owl:onProperty :methodName ;
        owl:someValuesFrom :notEmptyToken ] .

:mthd a owl:ObjectProperty, owl:FunctionalProperty ;
    rdfs:label "method"@en ;
    rdfs:comment "The HTTP method used for the HTTP request."@en ;
    rdfs:domain :Request ;
    rdfs:range :Method .

:methodName a owl:DatatypeProperty, owl:FunctionalProperty ;
    rdfs:label "method name"@en ;
    rdfs:comment "The HTTP method name used for the HTTP request."@en ;
    rdfs:domain :Method ;
    rdfs:range :notEmptyToken .

:notEmptyToken a rdfs:Datatype ;
    rdfs:label "Non-empty token"@en ;
    rdfs:comment "A token with at least one character" ;
    owl:equivalentClass [
        a rdfs:Datatype ;
        owl:onDatatype xsd:token ;
        owl:withRestrictions ([ xsd:minLength 1])] .

mthd:GET a :Method ;
    rdfs:label "GET" ;
    rdfs:isDefinedBy <http://tools.ietf.org/rfc/rfc7231#section-4.3.1> ;
    :methodName "GET" .

mthd:HEAD a :Method  ;
    rdfs:label "HEAD" ;
    rdfs:isDefinedBy <http://tools.ietf.org/rfc/rfc7231#section-4.3.2> ;
    :methodName "HEAD" .

mthd:POST a :Method  ;
    rdfs:label "POST" ;
    rdfs:isDefinedBy <http://tools.ietf.org/rfc/rfc7231#section-4.3.3> ;
    :methodName "POST" .

mthd:PUT a :Method  ;
    rdfs:label "PUT" ;
    rdfs:isDefinedBy <http://tools.ietf.org/rfc/rfc7231#section-4.3.4> ;
    :methodName "PUT" .

mthd:DELETE a :Method  ;
    rdfs:label "DELETE" ;
    rdfs:isDefinedBy <http://tools.ietf.org/rfc/rfc7231#section-4.3.5> ;
    :methodName "DELETE" .

mthd:CONNECT a :Method  ;
    rdfs:label "CONNECT" ;
    rdfs:isDefinedBy <http://tools.ietf.org/rfc/rfc7231#section-4.3.6> ;
    :methodName "CONNECT" .

mthd:OPTIONS a :Method  ;
    rdfs:label "OPTIONS" ;
    rdfs:isDefinedBy <http://tools.ietf.org/rfc/rfc7231#section-4.3.7> ;
    :methodName "OPTIONS" .

mthd:TRACE a :Method  ;
    rdfs:label "TRACE" ;
    rdfs:isDefinedBy <http://tools.ietf.org/rfc/rfc7231#section-4.3.8> ;
    :methodName "TRACE" .

mthd:PATCH a :Method  ;
    rdfs:label "PATCH" ;
    rdfs:isDefinedBy <http://tools.ietf.org/rfc/rfc5789> ;
    :methodName "PATCH" .

## ---- ##
## URI. ##
## ---- ##

:uri a owl:ObjectProperty, owl:FunctionalProperty ;
    rdfs:label "uri" ;
    rdfs:comment "Effective request URI" ;
    rdfs:domain :Request ;
    rdfs:range :URI .

:URI a owl:Class ;
    rdfs:label "URI description" ;
    rdfs:comment
    "A semantic description of the syntactic parts composing a URI."@en .

:scheme a owl:DatatypeProperty, owl:FunctionalProperty ;
    rdfs:label "scheme"@en ;
    rdfs:domain :URI ;
    rdfs:comment "The scheme part of an URI."@en .

:authority a owl:DatatypeProperty, owl:FunctionalProperty ;
    rdfs:label "authority"@en ;
    rdfs:domain :URI ;
    rdfs:comment "The authority part of an URI."@en .

:path a owl:DatatypeProperty, owl:FunctionalProperty ;
    rdfs:label "path"@en ;
    rdfs:domain :URI ;
    rdfs:comment "The path part of an URI."@en .

:query a owl:DatatypeProperty, owl:FunctionalProperty ;
    rdfs:label "query"@en ;
    rdfs:domain :URI ;
    rdfs:comment "The query part of an URI."@en .

:fragment a owl:DatatypeProperty, owl:FunctionalProperty ;
    rdfs:label "fragment"@en ;
    rdfs:domain :URI ;
    rdfs:comment "The fragment part of an URI."@en .

:idRes a owl:DatatypeProperty, owl:FunctionalProperty ;
    rdfs:label "resource"@en ;
    rdfs:comment "Everything except the query part"@en ;
    rdfs:domain :URI .

## -------- ##
## Headers. ##
## -------- ##

:Header a owl:Class ;
    rdfs:label "Header"@en ;
    rdfs:comment "A header in an HTTP message."@en ;
    rdfs:subClassOf [
        a owl:Restriction ;
        owl:onProperty :hdrName ;
        owl:someValuesFrom rdfs:Literal
        ] , [
        a owl:Restriction ;
        owl:onProperty :hdrValue ;
        owl:someValuesFrom rdfs:Literal
        ] .

:hdrName a owl:DatatypeProperty, owl:FunctionalProperty ;
    rdfs:label "header name"@en ;
    rdfs:comment "The name of an HTTP message header."@en ;
    rdfs:domain :Header ;
    rdfs:range rdfs:Literal .

:hdrValue a owl:DatatypeProperty, owl:FunctionalProperty ;
    rdfs:label "header value"@en ;
    rdfs:comment "The value of an HTTP message header."@en ;
    rdfs:domain :Header ;
    rdfs:range rdfs:Literal .

:hdr a owl:ObjectProperty ;
    rdfs:label "header"@en ;
    rdfs:comment "The headers in an HTTP message."@en ;
    rdfs:domain :Message ;
    rdfs:range :Header .

### Location Header property

hds:isLocationHeader a owl:ObjectProperty, owl:ReflexiveProperty ;
    rdfs:label "location header?" ;
    rdfs:domain :Header ;
    rdfs:range :Header .

:link a owl:ObjectProperty, owl:FunctionalProperty .

hds:LocationHeader a owl:Class ;
    rdfs:subClassOf [
        a owl:Restriction ;
        owl:onProperty hds:isLocationHeader ;
        owl:hasSelf true
        ] , [
        a owl:Restriction ;
        owl:onProperty :link ;
        owl:someValuesFrom :URI ;
        ] ;
    owl:equivalentClass [
        owl:intersectionOf (:Header [
                a owl:Restriction ;
                owl:onProperty :hdrName ;
                owl:hasValue "Location" ;
                ])] .

hds:location a owl:ObjectProperty ;
    rdfs:label "location" ;
    rdfs:domain :Response ;
    rdfs:range :URI ;
    owl:propertyChainAxiom (:hdr hds:isLocationHeader :link) .

## ----------------- ##
## Query parameters. ##
## ----------------- ##

:QueryParam a owl:Class ;
    rdfs:comment "A parameter for a part of a header value."@en ;
    rdfs:label "Query Parameter"@en .

:paramName a owl:DatatypeProperty, owl:FunctionalProperty ;
    rdfs:label "parameter name"@en ;
    rdfs:comment "The name of a query parameter."@en ;
    rdfs:domain :QueryParam ;
    rdfs:range rdfs:Literal .

:paramValue a owl:DatatypeProperty, owl:FunctionalProperty ;
    rdfs:label "parameter value"@en ;
    rdfs:comment "The literal value of a query parameter."@en ;
    rdfs:domain :QueryParam ;
    rdfs:range rdfs:Literal .

:queryParams a owl:ObjectProperty ;
    rdfs:label "query parameters"@en ;
    rdfs:comment "The parameters found in the query string part of a URL."@en ;
    rdfs:domain :URI ;
    rdfs:range :QueryParam .

## -------- ##
## Content. ##
## -------- ##

cnt:Content a owl:Class ;
    rdfs:label "Content"@en ;
    rdfs:comment
    "Representation of a content which can associated to various formats."@en .

sd:Graph a rdfs:Class ;
    rdfs:label "Graph"@en ;
    rdfs:comment
    "An instance of sd:Graph represents the description of an RDF graph."@en .

cnt:about a owl:ObjectProperty ;
    rdfs:label "graph"@en ;
    rdfs:comment "A property associating an RDF content with its RDF graph"@en ;
    rdfs:domain cnt:ContentAsRDF ;
    rdfs:range sd:Graph .

cnt:ContentAsRDF a owl:Class ;
    rdfs:label "RDF Content"@en ;
    rdfs:comment "RDF Content embedded in the body of an HTTP message"@en ;
    rdfs:subClassOf cnt:Content ;
    owl:equivalentClass [
        a owl:Restriction ;
        owl:onProperty cnt:about ;
        owl:cardinality 1 ] .

:body a owl:ObjectProperty, owl:FunctionalProperty ;
    rdfs:label "body"@en ;
    rdfs:comment "The entity body of an HTTP message."@en ;
    rdfs:domain :Message ;
    rdfs:range cnt:Content .

## ------------ ##
## Status codes ##
## ------------ ##

:StatusCode a owl:Class ;
    rdfs:label "Status code"@en ;
    owl:equivalentClass [
        a owl:Restriction ;
        owl:onProperty :statusCodeNumber ;
        owl:someValuesFrom :threeDigit ;
        ] ;
    rdfs:isDefinedBy <http://tools.ietf.org/rfc/rfc7231#section-6> ;
    rdfs:comment "The status code of an HTTP response."@en .

:sc a owl:ObjectProperty, owl:FunctionalProperty ;
    rdfs:label "status code"@en ;
    rdfs:domain :Response ;
    rdfs:range :StatusCode ;
    rdfs:isDefinedBy <http://tools.ietf.org/rfc/rfc7231#section-6> ;
    rdfs:comment "The status code of an HTTP response."@en .

:threeDigit a rdfs:Datatype ;
    rdfs:label "3-digit integer"@en ;
    rdfs:comment "A positive integer consisting in three digit" ;
    owl:equivalentClass [
        a rdfs:Datatype ;
        owl:onDatatype xsd:nonNegativeInteger ;
        owl:withRestrictions ([ xsd:maxInclusive 999])] .

:statusCodeNumber a owl:DatatypeProperty, owl:FunctionalProperty ;
    rdfs:label "status code number"@en ;
    rdfs:domain :StatusCode ;
    rdfs:range :threeDigit ;
    rdfs:isDefinedBy <http://tools.ietf.org/rfc/rfc7231#section-6> ;
    rdfs:comment "The status code number."@en .

sc:Informational a owl:Class ;
    owl:equivalentClass [ owl:intersectionOf (:StatusCode [
                a owl:Restriction ;
                owl:onProperty :statusCodeNumber ;
                owl:someValuesFrom [
                    a rdfs:Datatype ;
                    owl:onDatatype xsd:integer ;
                    owl:withRestrictions ([ xsd:minInclusive 100] [ xsd:maxInclusive 199])]])
        ] ;
    rdfs:label "Informational"@en ;
    rdfs:comment "A status code starting with 1, denoting Status an informational response"@en .

sc:Successful a owl:Class ;
    owl:equivalentClass [ owl:intersectionOf (:StatusCode [
                a owl:Restriction ;
                owl:onProperty :statusCodeNumber ;
                owl:someValuesFrom [
                    a rdfs:Datatype ;
                    owl:onDatatype xsd:integer ;
                    owl:withRestrictions ([ xsd:minInclusive 200] [ xsd:maxInclusive 299])]])
        ] ;
    rdfs:label "Successful"@en ;
    rdfs:comment "A status code starting with 2, denoting a successful interaction"@en .

sc:Redirection a owl:Class ;
    owl:equivalentClass [ owl:intersectionOf (:StatusCode [
                a owl:Restriction ;
                owl:onProperty :statusCodeNumber ;
                owl:someValuesFrom [
                    a rdfs:Datatype ;
                    owl:onDatatype xsd:integer ;
                    owl:withRestrictions ([ xsd:minInclusive 300] [ xsd:maxInclusive 399])]])
        ] ;
    rdfs:label "Redirection"@en ;
    rdfs:comment "A status code starting with 3"@en .

sc:ClientError a owl:Class ;
    owl:equivalentClass [ owl:intersectionOf (:StatusCode [
                a owl:Restriction ;
                owl:onProperty :statusCodeNumber ;
                owl:someValuesFrom [
                    a rdfs:Datatype ;
                    owl:onDatatype xsd:integer ;
                    owl:withRestrictions ([ xsd:minInclusive 400] [ xsd:maxInclusive 499])]])
        ] ;
    rdfs:label "Client Error"@en ;
    rdfs:comment "A status code starting with 4"@en .

sc:ServerError a owl:Class ;
    owl:equivalentClass [ owl:intersectionOf (:StatusCode [
                a owl:Restriction ;
                owl:onProperty :statusCodeNumber ;
                owl:someValuesFrom [
                    a rdfs:Datatype ;
                    owl:onDatatype xsd:integer ;
                    owl:withRestrictions ([ xsd:minInclusive 500] [ xsd:maxInclusive 599])]])
        ] ;
    rdfs:label "Server Error"@en ;
    rdfs:comment "A status code starting with 5"@en .

## Entities

sc:Accepted a :StatusCode ;
    rdfs:label "Accepted"@en ;
    rdfs:isDefinedBy <http://tools.ietf.org/rfc/rfc7231> ;
    :statusCodeNumber 202 .

sc:BadGateway a :StatusCode ;
    rdfs:label "Bad Gateway"@en ;
    rdfs:isDefinedBy <http://tools.ietf.org/rfc/rfc7231> ;
    :statusCodeNumber 502 .

sc:BadRequest a :StatusCode ;
    rdfs:label "Bad Request"@en ;
    rdfs:isDefinedBy <http://tools.ietf.org/rfc/rfc7231> ;
    :statusCodeNumber 400 .

sc:Conflict a :StatusCode ;
    rdfs:label "Conflict"@en ;
    rdfs:isDefinedBy <http://tools.ietf.org/rfc/rfc7231> ;
    :statusCodeNumber 409 .

sc:Continue a :StatusCode ;
    rdfs:label "Continue"@en ;
    rdfs:isDefinedBy <http://tools.ietf.org/rfc/rfc7231> ;
    :statusCodeNumber 100 .

sc:Created a :StatusCode ;
    rdfs:label "Created"@en ;
    rdfs:isDefinedBy <http://tools.ietf.org/rfc/rfc7231> ;
    :statusCodeNumber 201 .

sc:ExpectationFailed a :StatusCode ;
    rdfs:label "Expectation Failed"@en ;
    rdfs:isDefinedBy <http://tools.ietf.org/rfc/rfc7231> ;
    :statusCodeNumber 417 .

sc:FailedDependency a :StatusCode ;
    rdfs:label "Failed Dependency"@en ;
    rdfs:isDefinedBy <http://www.ietf.org/rfc/rfc4918.txt> ;
    :statusCodeNumber 424 .

sc:Forbidden a :StatusCode ;
    rdfs:label "Forbidden"@en ;
    rdfs:isDefinedBy <http://tools.ietf.org/rfc/rfc7231> ;
    :statusCodeNumber 403 .

sc:Found a :StatusCode ;
    rdfs:label "Found"@en ;
    rdfs:isDefinedBy <http://tools.ietf.org/rfc/rfc7231> ;
    :statusCodeNumber 302 .

sc:GatewayTimeout a :StatusCode ;
    rdfs:label "Gateway Timeout"@en ;
    rdfs:isDefinedBy <http://tools.ietf.org/rfc/rfc7231> ;
    :statusCodeNumber 504 .

sc:Gone a :StatusCode ;
    rdfs:label "Gone"@en ;
    rdfs:isDefinedBy <http://tools.ietf.org/rfc/rfc7231> ;
    :statusCodeNumber 410 .

sc:HTTPVersionNotSupported a :StatusCode ;
    rdfs:label "HTTP Version Not Supported"@en ;
    rdfs:isDefinedBy <http://tools.ietf.org/rfc/rfc7231> ;
    :statusCodeNumber 505 .

sc:IMUsed a :StatusCode ;
    rdfs:label "IM Used"@en ;
    rdfs:isDefinedBy <http://www.ietf.org/rfc/rfc3229.txt> ;
    :statusCodeNumber 226 .

sc:InsufficientStorage a :StatusCode ;
    rdfs:label "Insufficient Storage"@en ;
    rdfs:isDefinedBy <http://www.ietf.org/rfc/rfc4918.txt> ;
    :statusCodeNumber 507 .

sc:InternalServerError a :StatusCode ;
    rdfs:label "Internal Server Error"@en ;
    rdfs:isDefinedBy <http://tools.ietf.org/rfc/rfc7231> ;
    :statusCodeNumber 500 .

sc:LengthRequired a :StatusCode ;
    rdfs:label "Length Required"@en ;
    rdfs:isDefinedBy <http://tools.ietf.org/rfc/rfc7231> ;
    :statusCodeNumber 411 .

sc:Locked a :StatusCode ;
    rdfs:label "Locked"@en ;
    rdfs:isDefinedBy <http://www.ietf.org/rfc/rfc4918.txt> ;
    :statusCodeNumber 423 .

sc:MethodNotAllowed a :StatusCode ;
    rdfs:label "Method Not Allowed"@en ;
    rdfs:isDefinedBy <http://tools.ietf.org/rfc/rfc7231> ;
    :statusCodeNumber 405 .

sc:MovedPermanently a :StatusCode ;
    rdfs:label "Moved Permanently"@en ;
    rdfs:isDefinedBy <http://tools.ietf.org/rfc/rfc7231> ;
    :statusCodeNumber 301 .

sc:MultiStatus a :StatusCode ;
    rdfs:label "Multi-Status"@en ;
    rdfs:isDefinedBy <http://www.ietf.org/rfc/rfc4918.txt> ;
    :statusCodeNumber 207 .

sc:MultipleChoices a :StatusCode ;
    rdfs:label "Multiple Choices"@en ;
    rdfs:isDefinedBy <http://tools.ietf.org/rfc/rfc7231> ;
    :statusCodeNumber 300 .

sc:NoContent a :StatusCode ;
    rdfs:label "No Content"@en ;
    rdfs:isDefinedBy <http://tools.ietf.org/rfc/rfc7231> ;
    :statusCodeNumber 204 .

sc:NonAuthoritativeInformation a :StatusCode ;
    rdfs:label "Non-Authoritative Information"@en ;
    rdfs:isDefinedBy <http://tools.ietf.org/rfc/rfc7231> ;
    :statusCodeNumber 203 .

sc:NotAcceptable a :StatusCode ;
    rdfs:label "Not Acceptable"@en ;
    rdfs:isDefinedBy <http://tools.ietf.org/rfc/rfc7231> ;
    :statusCodeNumber 406 .

sc:NotExtended a :StatusCode ;
    rdfs:label "Not Extended"@en ;
    rdfs:isDefinedBy <http://www.ietf.org/rfc/rfc2774.txt> ;
    :statusCodeNumber 510 .

sc:NotFound a :StatusCode ;
    rdfs:label "Not Found"@en ;
    rdfs:isDefinedBy <http://tools.ietf.org/rfc/rfc7231> ;
    :statusCodeNumber 404 .

sc:NotImplemented a :StatusCode ;
    rdfs:label "Not Implemented"@en ;
    rdfs:isDefinedBy <http://tools.ietf.org/rfc/rfc7231> ;
    :statusCodeNumber 501 .

sc:NotModified a :StatusCode ;
    rdfs:label "Not Modified"@en ;
    rdfs:isDefinedBy <http://tools.ietf.org/rfc/rfc7231> ;
    :statusCodeNumber 304 .

sc:OK a :StatusCode ;
    rdfs:label "OK"@en ;
    rdfs:isDefinedBy <http://tools.ietf.org/rfc/rfc7231> ;
    :statusCodeNumber 200 .

sc:PartialContent a :StatusCode ;
    rdfs:label "Partial Content"@en ;
    rdfs:isDefinedBy <http://tools.ietf.org/rfc/rfc7231> ;
    :statusCodeNumber 206 .

sc:PaymentRequired a :StatusCode ;
    rdfs:label "Payment Required"@en ;
    rdfs:isDefinedBy <http://tools.ietf.org/rfc/rfc7231> ;
    :statusCodeNumber 402 .

sc:PreconditionFailed a :StatusCode ;
    rdfs:label "Precondition Failed"@en ;
    rdfs:isDefinedBy <http://tools.ietf.org/rfc/rfc7231> ;
    :statusCodeNumber 412 .

sc:Processing a :StatusCode ;
    rdfs:label "Processing"@en ;
    rdfs:isDefinedBy <http://www.ietf.org/rfc/rfc2518.txt> ;
    :statusCodeNumber 102 .

sc:ProxyAuthenticationRequired a :StatusCode ;
    rdfs:label "Proxy Authentication Required"@en ;
    rdfs:isDefinedBy <http://tools.ietf.org/rfc/rfc7231> ;
    :statusCodeNumber 407 .

sc:RequestEntityTooLarge a :StatusCode ;
    rdfs:label "Request Entity Too Large"@en ;
    rdfs:isDefinedBy <http://tools.ietf.org/rfc/rfc7231> ;
    :statusCodeNumber 413 .

sc:RequestTimeout a :StatusCode ;
    rdfs:label "Request Timeout"@en ;
    rdfs:isDefinedBy <http://tools.ietf.org/rfc/rfc7231> ;
    :statusCodeNumber 408 .

sc:RequestURITooLong a :StatusCode ;
    rdfs:label "Request-URI Too Long"@en ;
    rdfs:isDefinedBy <http://tools.ietf.org/rfc/rfc7231> ;
    :statusCodeNumber 414 .

sc:RequestedRangeNotSatisfiable a :StatusCode ;
    rdfs:label "Requested Range Not Satisfiable"@en ;
    rdfs:isDefinedBy <http://tools.ietf.org/rfc/rfc7231> ;
    :statusCodeNumber 416 .

sc:Reserved a :StatusCode ;
    rdfs:label "(Reserved)"@en ;
    rdfs:isDefinedBy <http://tools.ietf.org/rfc/rfc7231> ;
    :statusCodeNumber 306 .

sc:ResetContent a :StatusCode ;
    rdfs:label "Reset Content"@en ;
    rdfs:isDefinedBy <http://tools.ietf.org/rfc/rfc7231> ;
    :statusCodeNumber 205 .

sc:SeeOther a :StatusCode ;
    rdfs:label "See Other"@en ;
    rdfs:isDefinedBy <http://tools.ietf.org/rfc/rfc7231> ;
    :statusCodeNumber 303 .

sc:ServiceUnavailable a :StatusCode ;
    rdfs:label "Service Unavailable"@en ;
    rdfs:isDefinedBy <http://tools.ietf.org/rfc/rfc7231> ;
    :statusCodeNumber 503 .

sc:SwitchingProtocols a :StatusCode ;
    rdfs:label "Switching Protocols"@en ;
    rdfs:isDefinedBy <http://tools.ietf.org/rfc/rfc7231> ;
    :statusCodeNumber 101 .

sc:TemporaryRedirect a :StatusCode ;
    rdfs:label "Temporary Redirect"@en ;
    rdfs:isDefinedBy <http://tools.ietf.org/rfc/rfc7231> ;
    :statusCodeNumber 307 .

sc:Unauthorized a :StatusCode ;
    rdfs:label "Unauthorized"@en ;
    rdfs:isDefinedBy <http://tools.ietf.org/rfc/rfc7231> ;
    :statusCodeNumber 401 .

sc:UnprocessableEntity a :StatusCode ;
    rdfs:label "Unprocessable Entity"@en ;
    rdfs:isDefinedBy <http://www.ietf.org/rfc/rfc4918.txt> ;
    :statusCodeNumber 422 .

sc:UnsupportedMediaType a :StatusCode ;
    rdfs:label "Unsupported Media Type"@en ;
    rdfs:isDefinedBy <http://tools.ietf.org/rfc/rfc7231> ;
    :statusCodeNumber 415 .

sc:UpgradeRequired a :StatusCode ;
    rdfs:label "Upgrade Required"@en ;
    rdfs:isDefinedBy <http://www.ietf.org/rfc/rfc2817.txt> ;
    :statusCodeNumber 426 .

sc:UseProxy a :StatusCode ;
    rdfs:label "Use Proxy"@en ;
    rdfs:isDefinedBy <http://tools.ietf.org/rfc/rfc7231> ;
    :statusCodeNumber 305 .

sc:VariantAlsoNegotiates a :StatusCode ;
    rdfs:label "Variant Also Negotiates (Experimental)"@en ;
    rdfs:isDefinedBy <http://www.ietf.org/rfc/rfc2295.txt> ;
    :statusCodeNumber 506 .

## ----- ##
## Misc. ##
## ----- ##

:httpVersion a owl:DatatypeProperty ;
    rdfs:label "http version"@en ;
    rdfs:comment "The HTTP version of an HTTP message."@en ;
    rdfs:domain :Message ;
    rdfs:range rdfs:Literal ;
    rdfs:isDefinedBy <http://tools.ietf.org/rfc/rfc7231> .
\end{verbatim}

\end{document}